# Prime the search: Using large language models for guiding geometric task and motion planning by warm-starting tree search

Dongryung Lee*, Sejune Joo*, Kimin Lee and Beomjoon Kim

## Abstract

*The problem of relocating a set of objects to designated areas amidst movable obstacles can be framed as a Geometric Task and Motion Planning (G-TAMP), a subclass of task and motion planning problem (TAMP). Traditional approaches to G-TAMP have relied either on domain-independent heuristics or on learning from planning experience to guide the search, both of which typically demand significant computational resources or data. In contrast, humans often use common sense to intuitively decide which objects to manipulate in G-TAMP problems. Inspired by this, we propose leveraging Large Language Models (LLMs), which have common sense knowledge acquired from internet-scale data, to guide task planning in G-TAMP problems. To enable LLMs to perform geometric reasoning, we design a predicate-based prompt that encodes geometric information derived from a motion planning algorithm. We then query the LLM to generate a task plan, which is then used to search for a feasible set of continuous parameters. Since LLM is prone to mistakes, instead of committing to LLM's outputs we extend Monte Carlo Tree Search (MCTS) to a hybrid action space and use the LLM to guide the search. Unlike the previous approach that calls an LLM at every node and incurs high computational costs, we use it to warm-start the MCTS with the nodes explored in completing the LLM's task plan. On six different G-TAMP problems, we show our method outperforms previous LLM planners and pure search algorithms. Code can be found at https://github.com/iMSquared/prime-the-search.*

## 1. Introduction

Imagine a robot operating in a restaurant tasked with bringing items to a kitchen. Ideally, you would directly fetch target objects, but there often are obstacles in the way as shown in Figure 1. To solve the problem, the robot must figure out a sequence of objects and motions to clear obstacles and bring the goal objects to desired regions. These problems can be formulated as a geometric task and motion planning (G-TAMP) problem, a complex problem that involves hybrid action space that includes discrete actions, such as selecting which skill to use and which object to manipulate, as well as continuous actions, like determining the specific manipulation motion for each object. Additionally, the problem involves intricate reachability constraints among the movable obstacles.

Currently, there are two main approaches to G-TAMP problems. The first is pure-planning algorithms (Garrett et al., 2021), which typically integrate classical AI planning algorithms (Helmert, 2006) that use a domain-independent heuristic function with sampling or optimization to handle continuous parameters. While effective for general TAMP problems, this approach struggles with utilizing domain-*specific* information to identify the cause of infeasibility. For example, to determine whether the door needs to be opened in Figure 1, it must first attempt to plan a motion across the door, and only upon failure seek alternative discrete actions, that *may* open the door. This process

Graduate School of AI, Korea Advanced Institute of Science and Technology, Seoul, Korea

*co-first authors, equal contribution.

**Corresponding author:**
Beomjoon Kim, Korea Advanced Institute of Science and Technology, Graduate School of AI, 85, Hoegi-ro, Dongdaemun-gu, Seoul 02455, Korea.
Email: beomjoon.kim@kaist.ac.kr

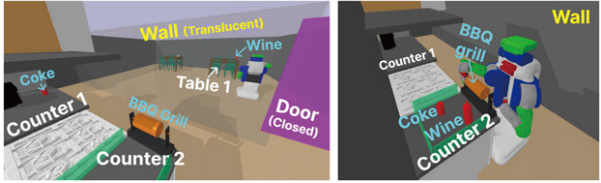

**Figure 1.** An example of a G-TAMP problem. Left: The initial configuration. The robot must bring the wine from Table 1, Coke from Counter 1, and a BBQ grill to Counter 2. The grill is already in its goal position but obstructing the placement of other goal objects, and a closed door between the kitchen and dining area must be opened to traverse these two areas. Right: A goal configuration. To achieve this, the robot must open the door, temporarily remove the grill from Counter 2, place other objects, and then bring back the grill.

is highly inefficient, often requiring numerous motion planning calls to identify the source of the failure. It would be much more efficient to perform causal reasoning such as "because the door is closed, and the door is in the way of moving the wine to Counter 2, we need to open the door."

Alternatively, we can use learning to guide search from planning experience. In particular, several works learn domain-specific heuristic functions or constraints to guide *task planning*, which is planning a sequence of discrete actions (Driess et al., 2020b; Kim et al., 2019; Kim and Shimanuki, 2019; Silver et al., 2021a). They have been shown to considerably improve planning speed compared to pure-planning strategies because they can learn through correlation. For example, in all successful plans, the door was opened in states where the robot had to move objects to the kitchen from the dining area, so we must open the door now. The problem, however, is that they typically require a significant amount of planning experience to acquire such knowledge, which is time-consuming to collect.

Our observation is that the knowledge required for task planning in a G-TAMP problem is straightforward for an agent with common sense, provided that the problem is expressed with the abstract representation that clearly encodes the constraints and goals. For example, in Figure 1, humans intuitively understand that the door needs to be opened if they know that the door is in the way of a manipulation motion. Based on this observation, we propose to use LLMs pre-trained on internet data for task planning, as they likely possess such common sense without additional training. The main challenge, however, lies in designing prompts that are effective across diverse problems and in managing erroneous outputs from an LLM caused by incorrect reasoning or hallucination (Kambhampati et al., 2024; Skreta et al., 2023).

One approach for designing the prompt is to adopt that of SayCan (Ahn et al., 2022), which consists of a task instruction, action history, and example plan. However, because it lacks state information, it is difficult to perform state-based causal reasoning. There also are several prompt designs that include predicate-based state information (Guan et al., 2023; Lin et al., 2023; Silver et al., 2024). However, they typically lack geometric information such as whether an obstacle is in the path to a particular object, which can only be evaluated via motion planning algorithms.

To solve this, we propose to use *geometric predicates*, which have shown to be effective in learning a relational value function for G-TAMP problems (Kim et al., 2022; Kim and Shimanuki, 2019), in our prompt to represent goals, states, and domains. As in previous work, we use a motion planning algorithm to compute these which encode reachability and occlusion. While our prompt can take various formats, we use the PDDL-style format for its clarity and proven efficacy with LLMs (Silver et al., 2024; Xie et al., 2023).

To combat the LLM's prediction errors (Lin et al., 2023; Skreta et al., 2023), we propose integrating an LLM with tree search, so that we can explore actions beyond those suggested by the LLM. The critical design choice here is how to structure the interface between the tree search and LLM. One approach is to invoke the LLM at every node. For example, LLM-MCTS (Zhao et al., 2024) combines an LLM with MCTS by deriving a policy from a batch of LLM responses and using it to determine which action to explore first. However, this method is highly inefficient since it calls the LLM at every node, and each LLM call involves processing a long sequence of tokens containing objects, states, and domain descriptions, with computation scaling quadratically to token length. Furthermore, MCTS is only restricted to discrete action spaces and is not applicable to G-TAMP.

To solve this, we first extend MCTS to hybrid action spaces and propose a method called **S**earch **T**ree **a**ugmented by **L**angauge **M**odel (STaLM), which uses an LLM to warm-start an MCTS. STaLM first queries the LLM for a batch of task plans that give discrete action choices but not continuous parameters. It concretizes these plans by searching for a feasible sequence of continuous parameters, and if this fails, initiates an MCTS that has been warm-started with the nodes that have been explored while trying to concretize the LLM's plans. Our intuition is that by concretizing batch queried task plans and using them for warm-started MCTS, the number of LLM queries can be minimized while leveraging the common sense from LLM to avoid numerous motion planning calls to identify the source of the failure. Figure 2 demonstrates our method.

In six different G-TAMP problems, we demonstrate that our prompt design is more effective than the existing prompt designs, and show that STaLM is more computationally efficient than the state-of-the-art pure-planning algorithms or other LLM-based planning algorithms.

## 2. Related work

### 2.1. Task and motion planning

Task and motion planning (TAMP) is a class of planning problems that require integrated discrete task planning and continuous motion planning (Garrett et al., 2021). One approach to TAMP is computing a task plan and refining it via sampling or optimization. Here, a task plan is a sequence of symbolic actions, which refer to an action with its continuous parameters unspecified. For example, PDDLStream (Garrett et al., 2020) creates a task plan with a classical planning system (Helmert, 2006) and refines the plan by sampling the continuous parameters with external functions called streams. eTAMP (Ren et al., 2021) uses tree search (Couëtoux et al., 2011) to sample continuous parameters for the task plans from top-k planner (Katz et al., 2020). Logic Geometric Programming (LGP) (Toussaint, 2015) frames TAMP as a continuous mathematical program and explicitly aims to optimize the final configuration represented in an objective function. LGP solvers use top-k-planner (Ortiz-Haro et al., 2022), tree search with hierarchical relaxation of constraints (Toussaint and Lopes, 2017), or tree search guided by hand-designed heuristics (Braun et al., 2021) to compute a task plan. They then solve continuous optimization problems with each symbolic action in the given task plan as a constraint to obtain continuous parameters. However, these methods lack common sense and require performing a search to compute a plan even for trivial matters like "pick the occluding object," rendering them inefficient.

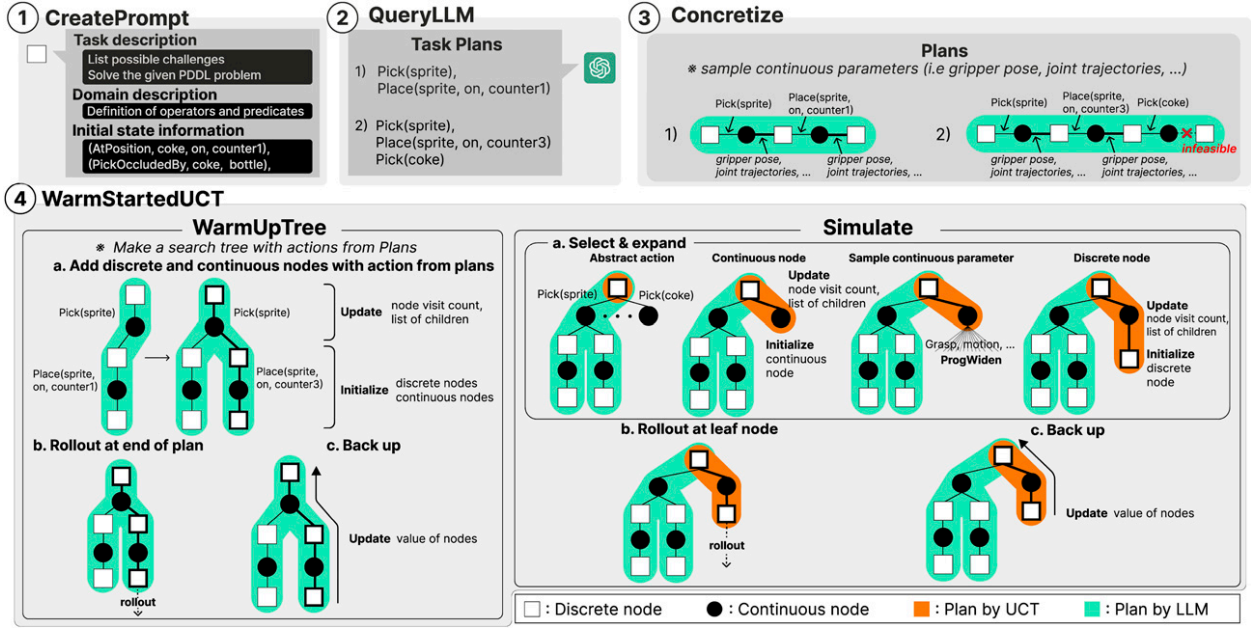

**Figure 2.** Overview of STaLM. (1) CreatePrompt uses a motion planning algorithm to compute state information for the prompt, such as one in Figure 3, for querying the LLM. (2) QueryLLM uses the prompt to generate $N_{batch}$ number of task plans, which are a sequence of discrete actions such as Pick (sprite), Place (sprite, on, counter1). (3) Concretize concretizes the given task plans by searching a feasible sequence of continuous parameters (e.g., gripper pose, joint trajectories, etc) for each discrete action. The squares denote discrete nodes for discrete action choices and circles denote continuous nodes for continuous parameter choices. After we have both discrete and continuous decisions, we simulate the next state by applying it and moving on to the next discrete node. If we succeed in finding a feasible concrete plan, we return the plan, otherwise, we commence the next step. (4) WarmStartedUCT consists of two processes: WarmUpTree and Simulate. WarmUpTree initializes the search tree with the states and actions explored in step (3), where the values of leaf nodes are estimated via rollout and backed-up along the tree. Simulate uses the usual MCTS operations on the warm-started tree: node selection using UCT, expansion, rollout, and backup. The orange bubble shows nodes explored by tree search, and the green bubble shows nodes explored using LLM's suggestions. We use Progressive Widening to expand at continuous nodes.

### 2.2. Learning to guide task and motion planning

There have been several attempts to adopt the intuitions from AlphaGo (Schrittwieser et al., 2019; Silver et al., 2016, 2017) to G-TAMP problems. Several studies propose to guide task planning by training a heuristic function based on images (Driess et al., 2020, 2020a), a graph that encodes the grounded predicates expressing the relationship among objects (Khodeir et al., 2021; Kim et al., 2022; Silver et al., 2021a), or context-dependent abstractions (Chitnis et al., 2021). While a raw image does not explicitly include reachability and occlusion, Kim et al. (Kim et al., 2022) directly use predicates about occlusions. Inspired by this, we also encode the state with a set of grounded geometric predicates. Other works train the samplers for continuous choices such as grasp or placement of the object using meta-learning (Chitnis et al., 2018) or generative models (Kim et al., 2018, 2020; Ortiz-Haro et al., 2021). However, to train any form of function, a significant computational cost is required to collect training data. In contrast, we use a pre-trained LLM to guide the search without additional learning. There are several studies (Kumar et al., 2023; Li and Silver, 2023; Silver et al., 2021b, 2022, 2023) that propose to learn predicates and operators, enabling more task-*specific* planning. Our work can be used in conjunction with these methods.

### 2.3. Planning with LLMs

Several methods (Ahn et al., 2022; Chen et al., 2023; Lin et al., 2023; Liu et al., 2023; Rana et al., 2023; Shinn et al., 2024; Silver et al., 2024; Singh et al., 2022; Zhao et al., 2024) use LLMs for planning with distinct skills, constraints, and objectives. Given a task instruction, SayCan (Ahn et al., 2022) outputs the probability of the next skill as a product of the LLM's probability of using that skill and skill affordance, the probability of that skill succeeding in the current state. Yet, its prompt lacks state information necessary for state-based causal reasoning. To enable this, a line of work (Guan et al., 2023; Lin et al., 2023; Liu et al., 2023; Silver et al., 2024; Singh et al., 2022) provides predicate-based state information to the LLM. However, even with state information, several works have shown that LLMs are not perfect at planning and tend to produce erroneous outputs (Kambhampati et al., 2024; Skreta et al., 2023). To combat this, other works (Shinn et al., 2024; Skreta et al., 2023) use LLM with self-refinement (Madaan et al., 2023) to generate a new response by providing past output and feedback. However, this strategy also shows limited improvement on domains that require a significant amount of diversity and exploration (Kambhampati et al., 2024; Shinn et al., 2024) since the choice of actions is fundamentally limited to actions given by the LLM. LLM-MCTS (Zhao et al., 2024) instead queries the LLM to obtain a policy to guide MCTS via PUCT (Rosin, 2011) and explore beyond actions that are suggested by LLM. Yet, LLM-MCTS has to call the LLM for every simulation, which is extremely expensive. In contrast, our framework only queries the LLM at the initial state.

## 3. Method

### 3.1. Problem formulation

We consider an environment that consists of a set of movable objects $\mathbf{O} = \{o_i\}_{i=1}^{n_O}$, a set of regions $\mathbf{R} = \{r_i\}_{i=1}^{n_W}$, and a set of doors $\mathbf{D} = \{d_i\}_{i=1}^{n_D}$. We model the world with a deterministic Markov Decision Process (MDP) with a state space $S$, a parameterized action space $\mathcal{A}$, a deterministic transition model $T$, and reward function $R$. A state is defined by the stable poses of movable objects, $\mathcal{P}_{o_i} \in SE(2)$, whether the doors are opened or closed $\mathcal{U}_{d_i} \in \{0, 1\}$, and robot configuration $c \in \mathcal{C}$, and is denoted as $s \in S$ where $s = (\mathcal{P}_{o_1}, ..., \mathcal{P}_{o_{n_O}}, \mathcal{U}_{d_1}, ..., \mathcal{U}_{d_{n_D}}, c)$. All entities have known and fixed shapes. An action $\alpha(\delta, \kappa) \in \mathcal{A}$ consists of an operator name $\alpha$, such as PLACE, a set discrete parameters $\delta$, such as a region to place an object down, and a set of continuous parameters $\kappa$, such as a trajectory.

Each $\alpha(\delta, \kappa)$ induces a mapping $T(s, \alpha(\delta, \kappa))$ from $s$ to $s' \in S$. If $\alpha(\delta, \kappa)$ cannot be legally executed at $s$, we let $s' = s$, absorbing the "failure" state and terminating the simulation. A goal $\mathcal{G}$ is given as a conjunction of (AtPosition$o_{goal}, dir, o_{ref}$), where $o_{goal} \in \mathbf{O}$, $dir \in \{$on, left, right, front, behind$\}$, and $o_{ref} \in \mathbf{O} \cup \mathbf{R}$. A problem instance consists of $(\mathbf{O}, \mathbf{R}, \mathbf{D}, s_0, \mathcal{G}, T, R, h)$, where $s_0$ is the initial state and $h$ is the planning horizon. The objective is to find a sequence of actions that satisfies $\mathcal{G}$.

We define an *action* as $\alpha(\delta, \kappa)$, which has a *discrete action* $\alpha(\delta)$ and *continuous parameter* $\kappa$, and a sequence of $\alpha(\delta)$ as a *task plan*. Our method consists of three main components: (1) a prompt design based on geometric predicates for querying the LLM for task plans, (2) a search algorithm that concretizes the task plan by finding a feasible set of continuous parameters, and (3) if unsuccessful, commences a hybrid action space Upper Confidence Tree (UCT) (Kocsis and Szepesvari, 2006) warm-started with explored nodes.

### 3.2. Predicate computation and prompt design

Our prompt consists of a task instruction, domain description, goal, objects, and the initial state as shown in Figure 3. We use a set of geometric predicates such as PICKOCCLUDEDBY and PLACEOCCLUDEDBY, and use motion planning algorithms to compute them. Concretely, we first compute the path to pick or place the designated object without considering other movable objects. Then, we check collisions between the swept volume of the motions and movable objects. If there is a collision, we set the occlusion predicate as true.

Part of our Task Instruction (Figure 3 top) asks for possible challenges for the problem. Empirically, without this, we have observed that LLMs often fail to respect an action's preconditions (e.g., pick the goal object despite being occluded by other objects). This was inspired by chain-of-thought (CoT) (Wei et al., 2022), which showed that LLM's reasoning capability improves when asked to generate intermediate reasoning steps. Our insight is that since our problem is geometric, in which occlusion relations are of main concern, this will make the LLM to respect the preconditions. An example prompt of STaLM is included in Appendix E.

### 3.3. Search Tree augmented by Language Model (STaLM)

---

**Algorithm 1:** STaLM ($s_0, h, N_{batch}, N_{budget}$)

---
1 **Global Variables:** $\mathcal{G}, H, \gamma$
2 TaskPlans ← QueryLLM(CreatePrompt($s_0, \mathcal{G}$), $N_{batch}$)
3 success, Plans ← Concretize(TaskPlans, $s_0, h$)
4 **if** success
5    **return** Plans
6 $Q$ ← WarmStartedUCT(Plans, $s_0, h, N_{budget}$).$Q$
7 **return** $\arg\max_{\mathfrak{a},\delta,\kappa} Q(s_0, \mathfrak{a}(\delta, \kappa))$

---

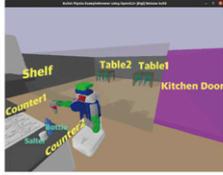

**Figure 3.** An example prompt for the state shown in the top-right corner (the image is not given to the LLM). Task instruction defines the output template and asks the LLM to state the challenges. Domain Description (Predicates) defines the set of predicates for our domain. Domain Description (Actions) gives operator definitions. The right column gives a set of entities, the initial state, and the goal defined by the predicates and the entities.

The key idea of STaLM is to use an LLM to perform task planning but complement it with additional search whenever the LLM fails to provide the solution. Unlike the previous work which queries the LLM at every time step, we use the states we explored during the concretization of task plans to warm up the tree search. This reduces the number of expensive LLM calls, but still directs the search into a promising region of the search space.

Algorithm 1 gives a pseudocode for STaLM. It takes an initial state $s_0$, planning horizon $h$, LLM query batch size $N_{batch}$, and planning budget $N_{budget}$ as inputs. The algorithm first queries the LLM for a batch of task plans with CreatePrompt, which creates our prompt from $s_0$ and $\mathcal{G}$ (L2), and concretizes the task plans via Concretize, which searches for a sequence of continuous parameters using sampling (L3). If any of the plans succeeds, we return the plan, otherwise, we commence WarmStartedUCT, an MCTS for hybrid action space. We pictorially explain how Algorithm 1 works in Figure 2.

---

**Algorithm 2:** Concretize(TaskPlans, $s$, $h$)

1 **Global Variables:** $T, H, \psi$
2 success $\leftarrow False$, Plans $\leftarrow []$
3 **for** TaskPlan *in* TaskPlans **do**
4 $\quad plan \leftarrow []$
5 $\quad$ **for** $\mathfrak{a}(\delta)$ *in* TaskPlan **do**
6 $\quad\quad \kappa \sim \psi(s, \mathfrak{a}(\delta))$, $s' \leftarrow T(s, \mathfrak{a}(\delta, \kappa))$
7 $\quad\quad$ plan.append($\mathfrak{a}(\delta, \kappa)$)
8 $\quad\quad$ **if** $h > H$ *or* $\mathfrak{a}(\delta, \kappa)$ is not feasible **then**
9 $\quad\quad\quad$ **break**
10 $\quad\quad$ **if** $s' \in \mathcal{G}$ **then**
11 $\quad\quad\quad$ success $\leftarrow True$
12 $\quad\quad\quad$ **break**
13 $\quad\quad s \leftarrow s', h \leftarrow h+1$
14 $\quad$ Plans.append(plan)
15 **return** success, Plans

Algorithm 2 shows the Concretize algorithm. We sample continuous parameters for the discrete actions and check whether the goal $\mathcal{G}$ can be satisfied with the discrete actions and sampled continuous parameters. Concretize takes as input TaskPlans, state $s$, and search depth $h$. For each discrete action $\mathfrak{a}(\delta)$ of TaskPlan, we sample continuous parameters $\kappa$ with continuous parameter sampler $\psi$ and run the transition model $T$ (L5-6). If the resulting action $\mathfrak{a}(\delta,\kappa)$ is infeasible or exceeds max search depth $H$, the iteration stops for the plan (L8-9). If the subsequent state $s'$ satisfies the goal $\mathcal{G}$, we terminate the process (L10-12). After iterating over all task plans, we return success and Plans (L15).

---

**Algorithm 3:** WarmStartedUCT(Plans, $s, h, N_{budget}$)

1 **Global Variables:** $T, R, H$
2　$\mathcal{T}(s) = \{Q(s,\cdot) = 0, n = 0, U = \emptyset, \}$
3　$\mathcal{T} \leftarrow$ WarmUpTree($\mathcal{T}, s,$ Plans)
4 **for** $i = 0$ to $N_{budget}$ **do**
5　　Simulate($s, h, 0, \mathcal{T}$)
6 **return** $\mathcal{T}$

---

If Concretize fails to find a feasible plan, WarmStartedUCT shown in Algorithm 3 uses the progress we made in concretizing the task plan given by LLM ("Plans" from L3 of Algorithm 1) to warm start the search tree for MCTS. WarmStartedUCT simply calls WarmUpTree to warm-up the search tree, and commences the search by calling Simulate for given number of budgets.

WarmStartedUCT starts by first initializing the search tree $\mathcal{T}$ with node value $Q$, number of visits $n$, and set of action parameters that have been tried $U$. To consider the hybrid action space, search tree $\mathcal{T}$ consists of two types of nodes: a discrete node $\mathcal{T}(s)$ where choices on discrete parameters of a discrete action $\mathfrak{a}(\delta)$ are made and a continuous node $\mathcal{T}(s_{\mathfrak{a}(\delta)})$ where choices on continuous parameters $\kappa$ for an action $\mathfrak{a}(\delta,\kappa)$ are made.

---

**Algorithm 4:** WarmUpTree($\mathcal{T}, s,$ Plans)

1 **Global Variables:** $T, R, H, \gamma$
2 **for** plan in Plans **do**
3　　AddToTree($\mathcal{T}, s, 0,$ plan)
4 **return** $\mathcal{T}$
5 **Function** AddToTree($\mathcal{T}, s, total,$ plan):
6　　$\mathfrak{a}(\delta,\kappa)$ = plan.pop()
　　/* Add Continuous Node to Tree */
7　　**if** $\mathfrak{a}(\delta) \notin \mathcal{T}(s).U$ **then**
8　　　$\mathcal{T}(s).U = \mathcal{T}(s).U \cup \{\mathfrak{a}(\delta)\}$
9　　　$\mathcal{T}(s_{\mathfrak{a}(\delta)}) = \{Q(s_{\mathfrak{a}(\delta)},\cdot) = 0, n = 0, U = \emptyset\}$
10　$\mathcal{T}(s_{\mathfrak{a}(\delta)}).n = \mathcal{T}(s_{\mathfrak{a}(\delta)}).n + 1$
　　/* Add Discrete Node to Tree */
11　**if** $\kappa \notin \mathcal{T}(s_{\mathfrak{a}(\delta)}).U$ **then**
12　　$\mathcal{T}(s_{\mathfrak{a}(\delta)}).U = \mathcal{T}(s_{\mathfrak{a}(\delta)}).U \cup \{\kappa\}$
13　　$s' \leftarrow T(s, \mathfrak{a}(\delta,\kappa))$
14　　$\mathcal{T}(s') = \{Q(s',\cdot) = 0, n = 0, U = \emptyset\}$
15　$s', r \leftarrow T(s, \mathfrak{a}(\delta,\kappa)), R(s, \mathfrak{a}(\delta,\kappa))$
16　$\mathcal{T}(s').n = \mathcal{T}(s').n + 1$
17　**if** len(plan)!$= 0$ **then**
18　　$total \leftarrow r + \gamma$ AddToTree($\mathcal{T}, s', total,$ plan)
19　**else**
20　　$total \leftarrow$ Rollout($s'$)
21　$\mathcal{T}(s_{\mathfrak{a}(\delta)}).Q(s_{\mathfrak{a}(\delta)},\kappa) \mathrel{+}= \frac{total - \mathcal{T}(s_{\mathfrak{a}(\delta)}).Q(s_{\mathfrak{a}(\delta)},\kappa)}{\mathcal{T}(s_{\mathfrak{a}(\delta)}).n}$
22　$\mathcal{T}(s).Q(s,\mathfrak{a}(\delta)) \mathrel{+}= \frac{total - \mathcal{T}(s).Q(s,\mathfrak{a}(\delta))}{\mathcal{T}(s).n}$
23　**return** $total$

---

Then, using the "Plans" from L3 of Algorithm 1, WarmUpTree algorithm described in Algorithm 4 adds discrete and continuous nodes to the search tree using actions from plans by LLM. WarmUpTree takes as input search tree $\mathcal{T}$, state $s$, and Plans and adds each plan to the search tree using AddToTree (L2-3). AddToTree takes as input search tree $\mathcal{T}$, state $s$, accumulated reward $total$ and plan. It pops an action $\mathfrak{a}(\delta,\kappa)$ from plan and initializes a continuous node $\mathcal{T}(s_{\mathfrak{a}(\delta)})$ with the discrete action $\mathfrak{a}(\delta)$ (L7-9). Similarly, it expands the continuous node with continuous parameter $\kappa$ from action $\mathfrak{a}(\delta,\kappa)$ and initializes a discrete node (L11-15). AddToTree is recursively called up to the last action of a plan (L17). The value of leaf node is estimated by Rollout and backed-up (L19-22).

---

**Algorithm 5:** Simulate($s, h, total, \mathcal{T}$)

1 **Global Variables:** $T, R, \psi, H, \gamma, k_\alpha, c_\alpha$
　/* Select discrete action with UCT */
2　$\hat{\mathfrak{a}}(\delta) \leftarrow \arg\max_{\mathfrak{a}(\delta)} \mathcal{T}(s).Q(s, \mathfrak{a}(\delta)) + c\sqrt{\frac{\log \mathcal{T}(s).n}{1+\mathcal{T}(s).child(\mathfrak{a}(\delta)).n}}$
3 **if** $\hat{\mathfrak{a}}(\delta) \notin \mathcal{T}(s).U$ **then**
4　　$\mathcal{T}(s).U = \mathcal{T}(s).U \cup \{\hat{\mathfrak{a}}(\delta)\}$
　　/* Add continuous node to tree */
5　　$\mathcal{T}(s_{\hat{\mathfrak{a}}(\delta)}) = \{Q(s_{\hat{\mathfrak{a}}(\delta)},\cdot) = 0, n = 0, U = \emptyset\}$
6　$\mathcal{T}(s_{\hat{\mathfrak{a}}(\delta)}).n = \mathcal{T}(s_{\hat{\mathfrak{a}}(\delta)}).n + 1$
　/* Progessive Widening */
7 **if** $|\mathcal{T}(s_{\hat{\mathfrak{a}}(\delta)}).U| \leq k_a(\mathcal{T}(s_{\hat{\mathfrak{a}}(\delta)}).n)^{c_\alpha}$ **then**
8　　$\kappa \sim \psi(s, \hat{\mathfrak{a}}(\delta)),\ \mathcal{T}(s_{\hat{\mathfrak{a}}(\delta)}).U = \mathcal{T}(s_{\hat{\mathfrak{a}}(\delta)}).U \cup \{\kappa\}$
　/* Select continuous parameter with UCT */
9　$\kappa \leftarrow \arg\max_{\kappa \in \mathcal{T}(s_{\hat{\mathfrak{a}}(\delta)}).U} \mathcal{T}(s_{\hat{\mathfrak{a}}(\delta)}).Q(s_{\hat{\mathfrak{a}}(\delta)},\kappa) + c\sqrt{\frac{\log \mathcal{T}(s_{\hat{\mathfrak{a}}(\delta)}).n}{1+\mathcal{T}(s_{\hat{\mathfrak{a}}(\delta)}).child(\kappa).n}}$
10　$s', r \leftarrow T(s, \hat{\mathfrak{a}}(\delta,\kappa)), R(s, \hat{\mathfrak{a}}(\delta,\kappa))$
　/* Add discrete node to tree */
11 **if** $\kappa \notin \mathcal{T}(s_{\hat{\mathfrak{a}}(\delta)}).U$ **then**
12　$\mathcal{T}(s') = \{Q(s',\cdot) = 0, n = 0, U = \emptyset\}$
13 $h \leftarrow h + 1,\ \mathcal{T}(s').n = \mathcal{T}(s').n + 1$
14 **if** $\mathcal{T}(s').U\ !=\emptyset$ **then**
15　**if** $h > H$ or $s' \neq$ *feasible* or $s' \in \mathcal{G}$ **then**
16　　$total \leftarrow r$
17　**else**
18　　$total \leftarrow r + \gamma$ Simulate($s', h, total, \mathcal{T}$)
19 **else**
20　$total \leftarrow$ Rollout($s'$)
21 $\mathcal{T}(s_{\hat{\mathfrak{a}}(\delta)}).Q(s_{\hat{\mathfrak{a}}(\delta)},\kappa) \mathrel{+}= \frac{total - \mathcal{T}(s_{\hat{\mathfrak{a}}(\delta)}).Q(s_{\hat{\mathfrak{a}}(\delta)},\kappa)}{\mathcal{T}(s_{\hat{\mathfrak{a}}(\delta)}).n}$
22 $\mathcal{T}(s).Q(s,\mathfrak{a}(\delta)) \mathrel{+}= \frac{total - \mathcal{T}(s).Q(s,\mathfrak{a}(\delta))}{\mathcal{T}(s).n}$
23 **return** $total$

---

The warmed-up search tree from WarmUpTree is then used for search by calling Simulate in Algorithm 5 for $N_{budget}$ times. Simulate takes as input state $s$, search depth $h$, accumulated reward $total$ and search tree $\mathcal{T}$. It begins by selecting a discrete action according to UCT, where exploration constant *c* is a hyperparameter, and adds a continuous node to the tree if the discrete action has not been tried (L3-5). Then, we use Progressive Widening (PW) (Couëtoux et al., 2011) to sample a new continuous parameter $\kappa$ only if the number of children in that node is below $k_\alpha \cdot N(\tau)^{c_\alpha}$, where $k > 0$ and $c_\alpha \in (0, 1)$ are hyperparameters. We select $\kappa$ using UCT and run the transition and reward model with selected action $\hat{a}(\delta, \kappa)$ to sample subsequent state $s'$ (L7-10). This process is recursively iterated until the leaf node (L17-18) or simulation is terminated due to reaching the max planning horizon $H$, encountering infeasible action, or achieving $\mathcal{G}$ (L15-16). At the leaf node, the value of the node is estimated by Rollout and backed-up along selected nodes of the search tree (L19-22).

## 4. Experiments

### 4.1. Experiment setup

We implement our domains that consist of two areas, a kitchen and a hall, separated by a door that must be opened to navigate the areas in PyBullet (Coumans and Bai, 2016). Each area has regions on which movable objects can be placed. We have the following assumptions: (1) the robot only uses its right arm for manipulation. (2) We have a predefined robot base pose $(x_r, y_r, \psi_r) \in SE(2)$ for each region and door. (3) The robot grasps object $o$ using a pre-defined grasp position $(x_g^o, y_g^o, z_g^o)$, and (4) the orientation of $o$ in the robot frame, $\omega_o$, is fixed during placement.

We design six problems to test the capabilities of different algorithms as shown in Figure 4. There are $n(\mathbf{R}) + 4 \times n(\mathbf{O})$ possible placements in our domains where 4 represents the four possible directions, left, right, front, and behind. For P1-6, the number of possible placement locations is 31, 23, 50, 27, 21, and 27 respectively. We perform 50 trials for P1 to P6 with max search depth $h$ of 20 and a time budget of 300 seconds for P1, 2, 4 and 600 seconds for P3, 5, 6.

The robot has 3 operators: PICK, PLACE, and OPEN. Table 1 summarizes each operator's discrete and continuous parameters. The continuous parameters that consist of base trajectory $\tau_{nav}$, gripper position $p_g$ and orientation $\omega_g$, and arm motion $\tau_g$ for each operator are sampled by the following procedures.

- PICK
  1. Use a motion planner, Probabilistic Roadmap (PRM), to compute $\tau_{nav}$ from the current robot base pose to $\mathcal{P}_b$.
  2. Gripper position $p_g$ is given as the discrete parameter for PICK. With the base pose fixed at $\mathcal{P}_b$, uniform-randomly sample the gripper orientation $\omega_g$ from the pre-defined ranges of pitch $\phi \in [0, 45°]$ and yaw $\chi \in [-60°, 60°]$ of the gripper.
  3. Find an inverse kinematics (IK) solution for the gripper pose $(p_g, \omega_g)$. Compute $\tau_g$ by linearly interpolating from the current robot arm joint configuration to the IK solution.

**Figure 4.** Description for problems showing the initial states and the goals. The purpose and possible challenges of each problem are listed. Goal objects are shown in cyan.

- OPEN
  1. Use a motion planner (PRM) to compute $\tau_{nav}$ from the current robot base pose to $\mathcal{P}_b$.
  2. Both $p_g$ and $\omega_g$ are given as the discrete parameters for OPEN. Find an IK solution for $(p_g, \omega_g)$ and compute $\tau_g$ by linear interpolation.
- PLACE
  1. Use a motion planner (PRM) to compute $\tau_{nav}$ from the current robot base pose to $\mathcal{P}_b$.
  2. $\omega_g$ is given as a discrete parameter for PLACE. With the base pose fixed at $\mathcal{P}_b$, uniform-randomly sample $p_g = (x_p, y_p)$ in the region for placement. We use rejection sampling until the direction of $p_g$ aligns with *dir* with respect to $o_{ref}$.
  3. Find an IK solution for $(p_g, \omega_g)$ and compute $\tau_g$ by linear interpolation.

If we fail to sample feasible $\tau_{nav}, p_g, \omega_g$, and $\tau_g$ within a fixed number of trials, we mark the action as infeasible. The goal is defined as a conjunction of (AtPosition $o_{goal}, dir, o_{ref}$). We give a reward of 3 for each AtPosition accomplished and $-6$ for sampling infeasible action. The discount factor $\gamma$ is 0.99, and if we find a successful plan, we stop the planning and execute it.

**Table 1.** Operator Descriptions for PICK, OPEN, and PLACE. A Dot Indicates an Unused Parameter. $O_{target}$ is the Target Object. *dir* and $O_{ref}$ are the Placement Direction and Reference Object Respectively. There is Only One $\mathcal{P}_b \in SE(2)$ for Each Target Region $r$. For PICK and OPEN, There is a Single Gripper Position $p_g$. In OPEN, We Have the Fixed Gripper Orientation $\omega_g$ for the Door. For PLACE, We Compute $\omega_g$ so That $O_{target}$'s Orientation with Respect to the Region on Which It is Placed Stays Same as Before the Pick. Both Base and Arm Motions, $\tau_{nav}$ and $\tau_g$ Respectively, are Computed Using Motion Planners and Apply to all Operators. For PLACE, $p_g = (x_p, y_p)$ is Randomly Sampled From the Placement Region's Surface. For PICK, $\omega_g$ Consists of Gripper's Pitch $\phi$ and Yaw $\chi$, Randomly Sampled From Specified Ranges.

|  | Discrete parameters | | | | | | Continuous parameters | | | |
| --- | --- | --- | --- | --- | --- | --- | --- | --- | --- | --- |
|  | $o_{target}$ | *dir* | $o_{ref}$ | $\mathcal{P}_b$ | $p_g$ | $\omega_g$ | $\tau_{nav}$ | $p_g$ | $\omega_g$ | $\tau_g$ |
| PICK | $o_{target} \in \mathbf{O}$ | . | . | $(x_r, y_r, \phi_r)$ | $(x_g^{o_{target}}, y_g^{o_{target}}, z_g^{o_{target}})$ | . | Motion planner | . | $(\phi, \chi)$ | Motion planner |
| OPEN | $o_{target} \in \mathbf{D}$ | . | . | $(x_r, y_r, \phi_r)$ | $(x_g^{o_{target}}, y_g^{o_{target}}, z_g^{o_{target}})$ | $\omega_g^{o_{target}}$ | . | . | . | . |
| PLACE | $o_{target} \in \mathbf{O}$ | On, left, right, front, behind | $o_{ref} \in \mathbf{O}$ if *dir* != on, $o_{ref} \in \mathbf{R}$ if *dir* == on | $(x_r, y_r, \phi_r)$ | . | $\omega_g^{o_{target}}$ | . | $(x_p, y_p)$ | . | . |

## 4.2. Baselines

We compare STaLM with the following baselines:

- *UCT* (Kocsis and Szepesvari, 2006): Standard UCT without LLM's guidance. $N_{budget}$ is set to 35.
- *UCT-with-Hcount* (Kim et al., 2022): Instead of rollout, we use a modified version of a hand-designed heuristic for G-TAMP, Hcount, as a value function for UCT. Hcount estimates the cost-to-go based on the number of occlusions for $o_{target}$. Details are given in the Appendix C. $N_{budget}$ is set to 35.
- *PDDLStream* (Garrett et al., 2020): a pure TAMP algorithm that uses a domain-independent heuristic function for guiding its search. We use an adaptive algorithm of PDDLStream.
- *EP-SayCan*: a variant of SayCan (Ahn et al., 2022). Because GPT does not provide token log probability of prompt required for the LLM score evaluation, we use the empirical policy distribution from (Zhao et al., 2024) with $N_{batch} = 5$ responses as a likelihood of action and use action's precondition as the affordance score. We follow the original prompt setting (task instruction and action history) of SayCan to examine the necessity of predicate computation and prompt design for G-TAMP problems.
- *Iterative-Replanning* (Shinn et al., 2024): uses our prompt to query the LLM for a single task plan and concretize it. If the plan does not succeed, Iterative-Replanning replans by appending up to two previously failed plans and calling the LLM for a new task plan. If no feasible plan is found after 5 attempts, the next action from the most recent plan is executed.
- *LLM-MCTS* (Zhao et al., 2024): computes the action distribution by counting the number of actions in the $N_{budget} = 5$ LLM responses and using it for action selection with PUCT (Rosin, 2011) in MCTS. $N_{budget}$ is 35.

STaLM uses $N_{batch} = 5$, $N_{budget} = 30$. For all LLM-based methods, we use `gpt-4-turbo-2024-04-09` with a decoding temperature of 1. Detailed hyperparameters of STaLM is included in Appendix B.

## 4.3. Results and analysis

Table 2 shows that STaLM outperforms all baselines in all problems in terms of planning speed and success rate. UCT performs poorly since it cannot simulate every possible placement for multiple steps ahead with $N_{budget} = 35$. UCT-with-Hcount calls the motion planner numerous times to compute the occlusion for every new state we encounter, leading to time-outs in *P1* and *P3-6*. STaLM, in contrast, computes occlusions only when querying the LLM. Furthermore, in *P5* and *P6*, UCT-with-Hcount fails to handle states in local optima where the objects already in the goal must be moved to another region because HCount explicitly penalizes such action. Such local optima are frequently encountered in *P5* when the tree search greedily follows the reward and is given as the initial state for *P6*.

PDDLStream treats each sample of continuous parameter as a "PDDL object" that exists in the world like a movable object or a region. Such PDDL objects are accumulated as the planning proceeds, slowing down the task planning (Vu et al., 2024). For instance, in P6, PDDLStream fails to sample collision-free placement pose for the coke occluded by the BBQ grill, but it can still reuse the gripper pose or base trajectories if the occluders are cleared. Therefore, numerous gripper poses and base trajectories that have been tried are stored in a cache, often exceeding 2000 in number. This results in about 100 seconds to make a task plan because the task planner must consider the combinations of accumulated PDDL objects, eventually leading to time-out.

SayCan avoids infeasible action via affordance score but cannot prioritize occlusion-clearing action since the prompt lacks state information and the LLM score of clearing the obstacle is indistinguishable from other actions. In contrast, our prompt includes state information about occlusions, so the LLM provides the task plan that clears occlusions.

**Table 2.** Success Rate and Average Planning Time of Success for STaLM and Baselines. Time-out *(t/o)* Indicates Cases Where Solutions Were Not Found Within the Max Time Limit. The bold font indicates the highest success rate for the given problem.

| Method | Metric | P1 | P2 | P3 | P4 | P5 | P6 |
| --- | --- | --- | --- | --- | --- | --- | --- |
| UCT | Success rate | 0.02 | 0.8 | 0.00 | 0.20 | 0.00 | 0.00 |
|  | Time (s) | 229.43 | 93.892 | t/o | 122.48 | t/o | t/o |
| UCT-with-Hcount | Success rate | 0.00 | 0.90 | 0.00 | 0.36 | 0.08 | 0.00 |
|  | Time (s) | t/o | 163.80 | t/o | 219.67 | 356.65 | t/o |
| PDDLStream | Success rate | 0.20 | 0.56 | 0.00 | 0.24 | 0.12 | 0.00 |
|  | Time (s) | 148.88 | 107.06 | t/o | 130.27 | 232.97 | t/o |
| EP-SayCan | Success rate | 0.02 | 0.14 | 0.00 | 0.00 | 0.00 | 0.00 |
|  | Time (s) | 54.01 | 81.68 | t/o | t/o | t/o | t/o |
| Iterative-Replanning | Success rate | 1.00 | 0.98 | 0.36 | 0.22 | 0.36 | 0.36 |
|  | Time (s) | 64.00 | 94.47 | 260.76 | 163.43 | 163.42 | 128.83 |
| LLM-MCTS | Success rate | 0.00 | 0.00 | 0.00 | 0.00 | 0.00 | 0.00 |
|  | Time (s) | t/o | t/o | t/o | t/o | t/o | t/o |
| STaLM (ours) | Success rate | **1.00** | **1.00** | **0.84** | **0.96** | **0.88** | **0.96** |
|  | Time (s) | 31.22 | 54.18 | 241.67 | 148.37 | 233.21 | 165.85 |

Iterative-replanning uses the LLM to modify a task plan based on the given feedback about what action of the task plan is infeasible. However for problems *P3-6*, where the set of feasible continuous parameters is relatively smaller than *P1-2*, a common failure mode was the LLM recklessly attributing the failure to the task plan even when the task plan could achieve success by further search of continuous parameters. STaLM, in contrast, is able to further search for continuous parameters by conducting MCTS.

All of LLM-MCTS leads to time-outs because, for each action, LLM-MCTS makes $N_{budget}$ number of LLM calls to compute an empirical action distribution where each call takes 10-20 seconds. STaLM instead makes a single LLM query to compute multiple plans, requiring fewer LLM calls for the same number of simulations and saving a significant amount of time. Example LLM responses of STaLM is shown in Appendix D.

We examine the effectiveness of integrating tree search with LLM by comparing with a variant of STaLM, STaLM without UCT that does not search further after Concretize. From Table 3, we see that STaLM without UCT shows lower performance than STaLM because it is limited to the LLM's response. We also evaluate STaLM and STaLM without UCT with a open-source model, Llama3.1-8B-Instruct. This model performs worse than STaLM with GPT-4 because larger models are better at step-by-step reasoning and instruction following (Zhao et al., 2023).

Computing occlusion predicates and asking LLM for challenges bring a non-trivial amount of increase in time to obtain task plans from LLM. So, to investigate the efficacy of our prompt design we test ablated versions of the prompt, as shown in Table 4. Without occlusion predicates, the LLM cannot perform state-based causal reasoning using geometric predicates, leading to the lowest performance of all ablated versions. Without asking for the challenge, the LLM returns responses that do not respect preconditions.

### 4.4. What kind of LLM error does STaLM fix?

Despite its common sense knowledge, LLMs can produce erroneous plans that fail to accomplish given goals (Kambhampati et al., 2024; Skreta et al., 2023). Such failures require re-querying the LLM and incur additional computation time. To evaluate STaLM's robustness to errors from LLMs, we analyze task plans generated by the LLM that failed to meet goal $\mathcal{G}$ during Concretize for problems P1 to 6. Figure 5 shows these failures, which fall into three categories: precondition violation, lack of 3D spatial understanding, and partially achieved goals.

"Precondition violation" is when the LLM-generated task plan is infeasible because a precondition of an action in the plan is not satisfied. This failure arises when the model hallucinates responses that deviate from the input context (Zhang et al., 2023). Examples include task plans that start with actions that are infeasible due occlusion or consecutive actions that violate preconditions described in the prompt.

"Lack of 3D spatial understanding" is when the LLM task plan is found to be infeasible during Concretize, which arises because natural language alone has inherent limitations in conveying the precise 3D object state information. For example, as shown in Figure 6, the grill is too large to be placed inside the green wall in *P5* and *P6*, so other objects must be placed on the counter first and then the grill should be moved at last. However, it is difficult to deliver the shapes of grill and the walls in a language prompt. LLM hence fails to understand that the grill must be placed at last and often generates the plan that moves the grill to the counter first, preventing other objects to be placed on the goal.

"Partially achieved goals" refers to cases where feasible continuous parameters are sampled for all actions, but the plan still fails to achieve goal $\mathcal{G}$. This failure happens when the LLM provides an incomplete response that fails to fully cover all aspects of a query for tasks that require sequential information, like step-by-step instructions (Luitel et al., 2023; Zhang et al., 2023). An example of this type of failure is a task plan that satisfies only four out of six goals in *P1*.

In STaLM, infeasible LLM task plans are not complete waste of computation. STaLM can reuse them by doing `WarmStartedUCT` on top of the failed plans and avoid the same failure, resulting in higher success rate. For instance, when the LLM overlooks preconditions and places the plate behind a closed door in *P4*, `WarmStartedUCT` finds that opening the door is more rewarding compared to the LLM task plan that immediately fails. Additionally, the LLM often tries to place the grill to the goal first in *P5* and *P6* because of the lack of 3D spatial understanding. However, STaLM recycles the unsuccessful LLM plan, avoid placing the grill to the goal, and finds a solution that puts the grill on other region by using `WarmStartedUCT`.

## 5. Limitations

One critical limitation of our work, or any work that utilizes LLM for TAMP problems, is that it is difficult to convey the geometric information of objects. As discussed in Section 4.4, describing the configuration of the objects in a 3D space in natural language is inherently limited. STaLM proposes using geometric predicates to mitigate this, yet its description is incomplete because the predicates only tell whether an object is occluded. For this reason, LLM often fails to be grounded in the low-level geometric state of the environment, leading to a planning failure due to a lack of 3D spatial understanding. However, there has been an attempt to encode the 3D shape into a latent representation and use it to perform geometric tasks like collision avoidance (Son et al., 2024). Therefore, incorporating such latent representation directly into the input to the LLM and finetuning the LLM can effectively convey the geometric information and better ground the LLM to the given geometric state.

Collision detection (CD) in the predicate computation is the bottleneck of UCT in STaLM, and its overhead increases quadratically with the number of objects. Many robotic simulators, including Pybullet (Coumans and Bai, 2016), adopt the Gilbert-Johnson-Keerthi (GJK) algorithm for CD. However, GJK requires convex object representations and due to limitations in handling arbitrary complex geometries, the computation becomes computationally expensive. Furthermore, the GJK algorithm involves branching that breaks the uniform computation flow, making it difficult to optimize with GPU-based compilers such as XLA (Sabne, 2020). However, we can speed up the CD algorithm by either simplifying the geometry with collection of primitive shapes (Sundaralingam et al., 2023) or replacing GJK algorithm with a neural networks (Son et al., 2024; Son and Kim, 2023). These advanced CD algorithms make not only a single collision check faster but also multiple collision checks parallelizable. With decreased overhead on the predicate computation, STaLM can be expanded to large-scale tasks.

**Table 3.** Success Rate of Ablated Version of STaLM. Specific Model Checkpoint for Llama-8B is `Llama3.1-8B-Instruct` and for GPT4 `gpt-4-turbo-2024-04-09`. The bold font indicates the highest success rate for the given problem.

| Model | Version | P1 | P2 | P3 | P4 | P5 | P6 |
|---|---|---|---|---|---|---|---|
| Llama-8B | w/o UCT | 0.96 | 0.64 | 0.36 | 0.12 | 0.20 | 0.00 |
| | STaLM | 0.98 | 0.80 | 0.46 | 0.16 | 0.32 | 0.34 |
| GPT4 | w/o UCT | 1.00 | 1.00 | 0.78 | 0.76 | 0.82 | 0.94 |
| | STaLM | **1.00** | **1.00** | **0.84** | **0.96** | **0.88** | **0.96** |

**Table 4.** Success Rate of Ablated Versions of Prompt Used by STaLM. The bold font indicates the highest success rate for the given problem.

| Prompting | | | | | | | |
|---|---|---|---|---|---|---|---|
| Occ. Info | Ask chall | P1 | P2 | P3 | P4 | P5 | P6 |
| - | - | 1.00 | 0.92 | 0.64 | 0.72 | 0.28 | 0.32 |
| - | ✓ | 0.98 | 0.84 | 0.64 | 0.50 | 0.20 | 0.20 |
| ✓ | - | 1.00 | 0.94 | 0.72 | 0.38 | 0.40 | 0.54 |
| ✓ | ✓ | **1.00** | **1.00** | **0.84** | **0.96** | **0.88** | **0.96** |

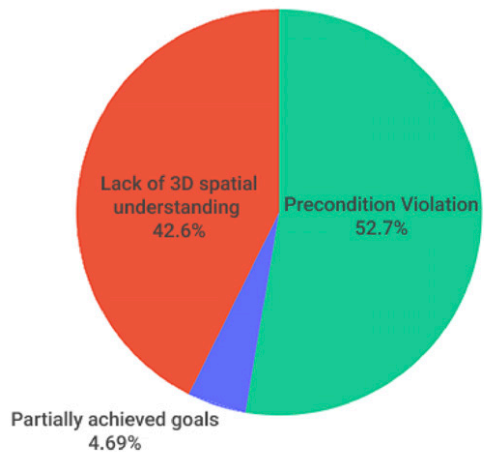
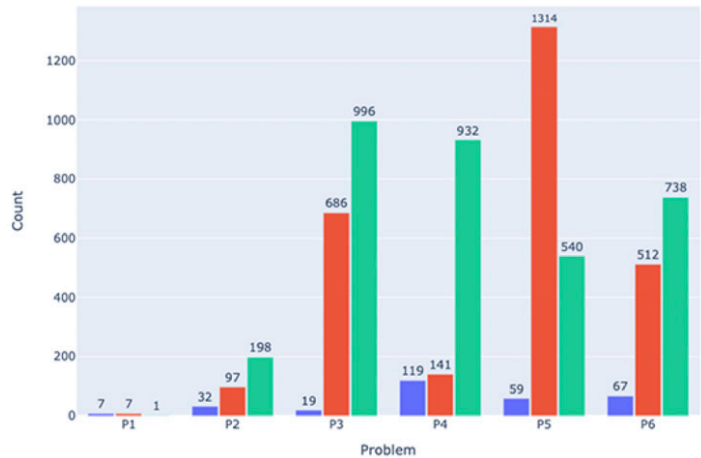

**Figure 5.** Analysis of LLM task plan failures while solving problems P1 to 6 with STaLM over 50 trials. We identify three types of failure cases: precondition violation, lack of 3D spatial understanding, and partially achieved goal. On the left, we provide the percentage of each failure type across all problems, and on the right, the count of each failure type for each individual problem.

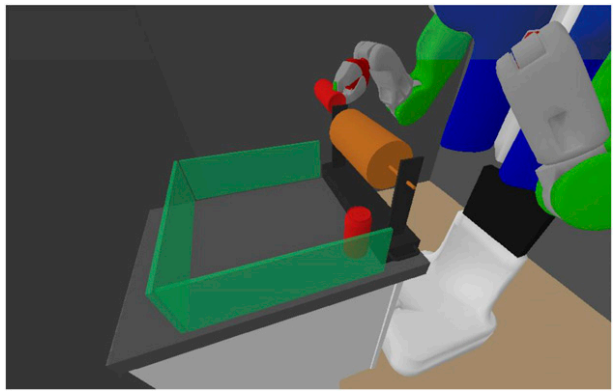

**Figure 6.** In *P5* and *P6*, the beef grill is too large to be placed inside the counter because there are green walls around the region. Therefore, if the grill is placed first, it prevents the robot from placing another object (red can in the hand) in the region. To accomplish the task, the robot must place all the other objects into the counter first and then place the beef grill at the front.

Another bottleneck is that STaLM does not guide the search for continuous choices. Although STaLM computes a potentially successful task plan, STaLM has to use extensive computation to find a set of continuous parameters, especially when the probability of sampling the feasible continuous parameters is small. To increase the sampling efficiency, several works (Ahn et al., 2023; Chitnis et al., 2016, 2018; Kim et al., 2018, 2020, 2022, 2022; Ortiz-Haro et al., 2021, 2022) suggest learning a sampler to focus the search on more promising regions based on past experience. Following their intuition, integrating an intelligent sampler is necessary to improve STaLM's efficiency regarding continuous choices. Recently, *Yang* et al. (Yang et al., 2023) adopted diffusion models to sample feasible object configurations conditioned on the language description. If this sampler is integrated into STaLM, the language will guide the search for both task plan and continuous parameters, making STaLM a more efficient bi-level planning algorithm.

Another important limitation of our work is that we assume full-observability of state with a deterministic transition model. To deploy this in the real world, we must also address the problem of state estimation and planning under the perception uncertainty. To account for the uncertainty, the environment should be modeled as partially observable Markov decision process (POMDP) (Garrett et al., 2019; Kaelbling and Lozano-Perez, 2013). Inspired by *Garrett* et al. (Garrett et al., 2019), we can integrate particle-filter based approach on STaLM's `Warm-StartedUCT` to plan under the uncertainty on current state estimation. In addition, with the advent of the foundation models with multi-modal inputs, some works delegate not only task planning but also state estimation to the foundation models (Hu et al., 2023). Similarly, we can fine-tune these models to output state estimation from the sensory input along with the task plan to account for the plan that is grounded on common sense as well as the perception.

## Declaration of conflicting interests

The author(s) declared no potential conflicts of interest with respect to the research, authorship, and/or publication of this article.

## Funding

The author(s) disclosed receipt of the following financial support for the research, authorship, and/or publication of this article: This work was supported by Institute of Information & communications Technology Planning & Evaluation (IITP) grant and National Research Foundation of Korea (NRF) funded by the Korea government (MSIT) (No.2019-0-00075, Artificial Intelligence Graduate School Program (KAIST)), (No.2022-0-00311, Development of Goal-Oriented Reinforcement Learning Techniques for Contact-Rich Robotic Manipulation of Everyday Objects), (No. 2022-0-00612, Geometric and Physical Commonsense Reasoning based Behavior Intelligence for Embodied AI), (No. RS-2024-00359085, Foundation model for learning-based humanoid robot that can understand and achieve language commands in unstructured human environments), (No. RS-2024-00509279, Global AI Frontier Lab).

## ORCID iDs

Dongryung Lee  https://orcid.org/0009-0006-1958-6883
Sejune Joo  https://orcid.org/0009-0009-1073-2321
Beomjoon Kim  https://orcid.org/0000-0002-8888-7253

## Supplemental Material

Supplemental Material for this article is available online.

## References

Ahn M, Brohan A, Brown N, et al. (2022) Do as i can, not as i say: grounding language in robotic affordances. In: Conference on Robot Learning, Seoul, Korea, September 27 - 30, 2025.

Ahn J, Son S, Lee D, et al. (2023) Preference learning for guiding the tree search in continuous pomdps. In: Conference on Robot Learning, Seoul, Korea, September 27 - 30, 2025.

Braun CV, Ortiz-Haro J, Toussaint M, et al. (2021) *Rhh-lgp: Receding Horizon and Heuristics-Based Logic-Geometric Programming for Task and Motion Planning*. Piscataway: 2022 IEEE/RSJ International Conference on Intelligent Robots and Systems (IROS).

Chen Y, Arkin J, Zhang Y, et al. (2023) *Autotamp: Autoregressive Task and Motion Planning with Llms as Translators and*

# Appendix

## A Caching state computations

We cache grounded predicates (literals) to reduce redundant computations. During PICK, all the objects are not moved, so we reuse all the literals. For OPEN, the occlusion by the door is cleared, so we reuse all the literals except those about occlusions by the door. PLACE removes an object from one region and adds it to another. Therefore, we discard any literals by the moved object and recompute the occlusion for the objects in the target region only. We also save intermediate computations. We cache gripper pose sampled during PICKOCCLUDEDBY and reuse them to grasp the same target for PLACEOCCLUDEDBY. The robot path computed for each object without movable objects stays the same within the same state, so we cache all the collisions and reuse them in the same state.

## B Hyperparameters of STaLM

For STaLM, we use $(N_{batch}, N_{budget}) = (5, 30)$ and `gpt-4-turbo-2024-04-09` with a decoding temperature of 1 to generate TaskPlans in `QueryLLM`. For WarmStartUCT of STaLM, We use UCT exploration constant $c = 50$, PW constants $(k_\alpha, c_\alpha) = (1.5, 0.15)$.

## C Hand-designed heuristics

The modified Hcount (Kim et al., 2022) used in UCT-with-Hcount is

$$Hcount = |M| - |O_{\texttt{InGoal}}| + \mathbb{1}_{o_{target} \in O_{\texttt{InGoal}}} \\ - \mathbb{1}_{o_{target} \notin O_{\texttt{InGoal}} \bigwedge o_{target} \in O_{goal}}$$

$M$ is the set of objects that occludes the goal objects $O_{goal}$ and that recursively occludes those occluders. $O_{\texttt{InGoal}}$ refers to the objects already in the goal. We add 1 if $o_{target} \in O_{\texttt{InGoal}}$ and discount 1 if $o_{target} \notin O_{\texttt{InGoal}}$ is an goal object. Since Hcount computes cost-to-go, we convert it to a value function for a node by using $Q = 3 \cdot n(O_{goal}) - 3 \cdot Hcount$, where $Q$ is the node value.

## D Example of LLM response in STaLM

In the following Figures 7–12, we provide examples of responses from LLM queries made in STaLM for *P1-6*.

## E Example of prompt used in STaLM

In Figure 13, we provide an example of a prompt used to generate task plans with LLM in *P5*.

---

```
## Possible Challenges for unachieved goals based on current state ##
1. Multiple objects are occluded by 'kitchen_door', which makes picking them difficult unless 'kitchen_door' is opened.
2. The locations of the objects 'salter1', 'salter2', 'bottle1', 'bottle2', and 'bottle3' need to be changed, but their placement may be occluded by various objects or by 'kitchen_door'.

## Plan ##
plan = [ ('open', 'kitchen_door'), # Open the kitchen door to prevent occlusion
('pick', 'bottle1'), # Pick bottle1 from table2
('place', 'bottle1', 'on', 'table1'), # Place bottle1 on table1
('pick', 'bottle2'), # Pick bottle2 from minifridge
('place', 'bottle2', 'on', 'counter2'), # Place bottle2 on counter2
('pick', 'salter1'), # Pick salter1 from counter1
('place', 'salter1', 'on', 'counter2'), # Place salter1 on counter2
('pick', 'salter2'), # Pick salter2 from counter2
('place', 'salter2', 'on', 'shelf'), # Place salter2 on shelf
('pick', 'bottle3'), # Pick bottle3 from table2
('place', 'bottle3', 'on', 'table1'), # Place bottle3 on table1
('pick', 'salter3'), # Pick salter3 from table1
('place', 'salter3', 'on', 'table2') # Place salter3 on table2
]
```

**Figure 7.** Example LLM response of STaLM in P1.

```
## Possible Challenges for unachieved goals based on current state ##
1. **Relocation of 'salter'**: The salter is currently positioned on 'counter2' and needs to be relocated to 'table2' as per the goal. However, any pick action for the salter is occluded by both 'bottle2' and 'bottle3'.
2. **Accessibility of 'salter'**: Before 'salter' can be picked up, the occluding items ('bottle2' and 'bottle3') need to be moved.
3. **Placement of 'salter' by 'bottle1' on 'table2'**: The specified final position for the 'salter' is 'right_of' 'bottle1' on 'table2'. Currently, 'bottle1' is already on 'table2'. So, the 'salter' needs to be placed adjacent to it correctly.

## Plan ##
plan = [('pick', 'bottle3'), # Pick up bottle3 which occludes picking up the salter
('place', 'bottle3', 'left _of', 'bottle2'), # Temporarily place it to clear the path
('pick', 'bottle2'), # Pick up bottle2 which also occludes the salter
('place', 'bottle2', 'left _of', 'bottle3'), # Temporarily place it so salter can be accessed
('pick', 'salter'), # Now pick up the salter
('place', 'salter', 'on', 'table2'), # Place the salter on table2
('place', 'salter', 'right _of', 'bottle1'), # Finally, place the salter right of bottle1 as required in the goal ]
```

**Figure 8.** Example LLM response of STaLM in P2.

## Possible Challenges for unachieved goals based on current state ##
1. **Moving objects to different positions**: To achieve the goal of having milk and coke on 'counter2', these objects need to be picked up and placed on the counter.
2. **Obstructed Picks**: The initial state mentions occlusions in picking certain items including milk and coke, thus complicating the pick action.
3. **Unavailable Target Region**: 'counter2' is not declared in the problem's object definitions, assuming it's an error, and considering placement on 'counter1' instead.

## Plan ##
plan = [('pick', 'coke'),
('place', 'coke', 'on', 'counter1'),
('pick', 'milk'),
('place', 'milk', 'on', 'counter1') ]

**Figure 9.** Example LLM response of STaLM in P3.

## Possible Challenges for unachieved goals based on current state ##
1. **Obstacles and Occlusions**: Moving or placing objects will be a significant challenge due to multiple occlusions, as detailed in the problem.
2. **Opening the Kitchen Door**: The kitchen door is currently closed ('Closed kitchen_door'), which is occluding several actions for moving objects. Thus, the door must be opened first to reduce occlusions and facilitate other actions.
3. **Handling Occluded Actions**: The picking up of 'salter' is currently occluded by both 'bottle2' and 'bottle3'. Similarly, placing 'plate' on 'counter1' has occlusions.
4. **Switch between picking and placing**: Robot will have to alternately pick and place objects, and this requires ensuring that the robot hand is available ('HandAvailable').

## Plan ##
plan = [ ('open', 'kitchen_door'), # Open the kitchen door to remove several occlusions
('pick', 'plate'), # Pick up the plate, assuming the kitchen_door being open has resolved the occlusion
('place', 'plate', 'on', 'counter1'), # Place the plate on counter1
('pick', 'salter'), # Pick up the salter
('place', 'salter', 'on', 'table2'), # Place the salter on table2
('place', 'salter', 'right_of', 'bottle1') # Place the salter to the right of bottle1 on table2]

**Figure 10.** Example LLM response of STaLM in P4.

## Possible Challenges for Unachieved Goals Based on Current State ##
1. The 'coke' and 'wine' need to be placed on 'counter2' but their current positions and potential placements are occluded by the 'beef_grill' and the 'kitchen_door'.
2. 'PickOccludedBy' predicates indicate that 'coke' and 'beef_grill' are occluded by the 'kitchen_door', which is closed.
3. 'AtPosition' predicates show that all movable objects are initially located on different counters and tables than 'counter2', where the objective is to position them.

To achieve the goal, certain actions involving moving occluding objects or opening doors will be needed.

## Plan ##
plan = [('open', 'kitchen_door'), # Open the kitchen door to ensure picking of coke and beef_grill is possible
('pick', 'coke'), # Pick coke, requirement: hand available and not unsafe to pick
('place', 'coke', 'on', 'counter2'), # Place coke on counter2, addressing one part of the goal
('pick', 'beef_grill'), # Pick beef_grill now that kitchen_door is open
('place', 'beef_grill', 'on', 'counter2'), # Place beef_grill on counter2, addressing another part of the goal
('pick', 'wine'), # Pick wine from its initial position
('place', 'wine', 'on', 'counter2') # Place wine on counter2; achieving the final part of the goal]

**Figure 11.** Example LLM response of STaLM in P5.

## Possible Challenges for Unachieved Goals Based on Current State ##
1. The 'coke' and 'wine' need to be placed on 'counter2' but their current positions and potential placements are occluded by the 'beef_grill' and the 'kitchen_door'.
2. 'PickOccludedBy' predicates indicate that 'coke' and 'beef_grill' are occluded by the 'kitchen_door', which is closed.
3. 'AtPosition' predicates show that all movable objects are initially located on different counters and tables than 'counter2', where the objective is to position them.

To achieve the goal, certain actions involving moving occluding objects or opening doors will be needed.

## Plan ##
plan = [('open', 'kitchen_door'), # Open the kitchen door to ensure picking of coke and beef_grill is possible
('pick', 'coke'), # Pick coke, requirement: hand available and not unsafe to pick
('place', 'coke', 'on', 'counter2'), # Place coke on counter2, addressing one part of the goal
('pick', 'beef_grill'), # Pick beef_grill now that kitchen_door is open
('place', 'beef_grill', 'on', 'counter2'), # Place beef_grill on counter2, addressing another part of the goal
('pick', 'wine'), # Pick wine from its initial position
('place', 'wine', 'on', 'counter2') # Place wine on counter2; achieving the final part of the goal ]

**Figure 12.** Example LLM response of STaLM in P6.

**System prompt**

You are an expert proficient in PDDL and planning actions for a problem. Your response should follow this template:
## Possible Challenges for unachieved goals based on current state ##

## Plan for unachieved goals ## plan = [('action_type', 'args_1', 'args_2',...), ]

**User prompt**

### Domain ###
(define (domain shop)
(:requirements :typing)
(:types movable_object region openable - object )
(:constants on left_of right_of front_of behind_of - direction )

(:predicates
(RobotHolding ?movable_object) ; True if robot is holding movable_object
(HandAvailable ) ; True if robot hand is available
(AtPosition ?subject ?direction ?reference) ; True if subject is at direction of reference
(IsClosed ?door) ; True if the door is closed
(PickOccludedBy ?subject ?occluder) ; True is action (pick, subject) is occluded by occluder
(PlaceOccludedBy ?subject ?direction ?reference ?occluder) ; True if action (place, subject, direction, reference) is occluded by occluder )

(:action pick ; example ('pick', 'bottle')
:parameters (?subject )
:precondition (and (HandAvailable ) (not (UnsafePick ?subject)) )
:effect (and (not (HandAvailable )) (RobotHolding ?subject) (not (AtPosition ?subject ?direction ?reference)) ) )

(:action place ; example ('place', 'bottle', 'behind_of', 'can')
:parameters (?subject ?direction ?reference)
:precondition (and (RobotHolding ?subject) (not (UnsafePlace ?subject ?direction ?reference)) )
:effect (and (not (RobotHolding )) (HandAvailable ) (AtPosition ?subject ?direction ?reference) ) )

(:action open ; example ('open', 'door')
:parameters (?subject )
:precondition (and (IsClosed ?subject) (HandAvailable ) )
:effect (and (not (IsClosed ?subject)) ) )

(:derived (UnsafePick ?subject)
(exists (?occluder) ((PickOccludedBy ?subject ?occluder))) )

(:derived (UnsafePlace ?subject ?direction ?reference)
(exists (?occluder) ((PlaceOccludedBy ?subject ?direction ?reference ?occluder))) )

### Problem ###
(:objects coke - movable_object sprite - movable_object mango_juice - movable_object dr_pepper - movable_object pork_grill - movable_object beef_grill - movable_object kitchen_door - openable counter1 - region counter2 - region counter3 - region counter5 - region )

(:init
[('HandAvailable', ''), ('AtPosition', 'coke', 'on', 'counter2'), ('AtPosition', 'beef_grill', 'on', 'counter2'), ('AtPosition', 'coke', 'behind_of', 'beef_grill'), ('AtPosition', 'beef_grill', 'front_of', 'coke'), ('AtPosition', 'sprite', 'on', 'counter3'), ('AtPosition', 'mango_juice', 'on', 'counter5'), ('AtPosition', 'dr_pepper', 'on', 'counter5'), ('AtPosition', 'mango_juice', 'behind_of', 'dr_pepper'), ('AtPosition', 'dr_pepper', 'front_of', 'mango_juice'), ('PickOccludedBy', 'coke', 'beef_grill'), ('PlaceOccludedBy', 'beef_grill', 'left_of', 'coke', 'coke'), ('PlaceOccludedBy', 'beef_grill', 'left_of', 'sprite', 'sprite'), ('PlaceOccludedBy', 'beef_grill', 'right_of', 'mango_juice', 'dr_pepper'), ('PlaceOccludedBy', 'beef_grill', 'right_of', 'mango_juice', 'mango_juice'), ('PlaceOccludedBy', 'beef_grill', 'left_of', 'dr_pepper', 'dr_pepper'), ('PlaceOccludedBy', 'sprite', 'on', 'counter2', 'beef_grill'), ('PlaceOccludedBy', 'sprite', 'left_of', 'coke', 'beef_grill'), ('PlaceOccludedBy', 'sprite', 'right_of', 'coke', 'coke'), ('PlaceOccludedBy', 'sprite', 'right_of', 'coke', 'beef_grill'), ('PlaceOccludedBy', 'sprite', 'behind_of', 'coke', 'beef_grill'), ('PlaceOccludedBy', 'sprite', 'behind_of', 'beef_grill', 'coke'), ('PlaceOccludedBy', 'sprite', 'behind_of', 'beef_grill', 'beef_grill'), ('PlaceOccludedBy', 'sprite', 'behind_of', 'mango_juice', 'dr_pepper'), ('PlaceOccludedBy', 'mango_juice', 'left_of', 'coke', 'beef_grill'), ('PlaceOccludedBy', 'mango_juice', 'right_of', 'coke', 'beef_grill'), ('PlaceOccludedBy', 'mango_juice', 'behind_of', 'coke', 'coke'), ('PlaceOccludedBy', 'mango_juice', 'behind_of', 'coke', 'beef_grill'), ('PlaceOccludedBy', 'mango_juice', 'behind_of', 'beef_grill', 'coke'), ('PlaceOccludedBy', 'mango_juice', 'behind_of', 'beef_grill', 'beef_grill'), ('PlaceOccludedBy', 'mango_juice', 'behind_of', 'sprite', 'sprite'), ('PlaceOccludedBy', 'mango_juice', 'front_of', 'dr_pepper', 'dr_pepper'), ('PlaceOccludedBy', 'dr_pepper', 'on', 'counter2', 'beef_grill'), ('PlaceOccludedBy', 'dr_pepper', 'left_of', 'coke', 'beef_grill'), ('PlaceOccludedBy', 'dr_pepper', 'right_of', 'coke', 'beef_grill'), ('PlaceOccludedBy', 'dr_pepper', 'behind_of', 'coke', 'coke'), ('PlaceOccludedBy', 'dr_pepper', 'behind_of', 'coke', 'beef_grill'), ('PlaceOccludedBy', 'dr_pepper', 'behind_of', 'beef_grill', 'coke'), ('PlaceOccludedBy', 'dr_pepper', 'behind_of', 'beef_grill', 'beef_grill')] )

(:goal (and (AtPosition coke on counter1) (AtPosition sprite on counter1) (AtPosition beef_grill on counter1) ) )

Generate a plan to achieve the goals from init.

**Figure 13.** Example of prompt used in STaLM in P5.